\documentclass[conference]{IEEEtran}
\IEEEoverridecommandlockouts

\usepackage{cite}
\usepackage{amsmath,amssymb,amsfonts}
\usepackage{algorithmic}
\usepackage{graphicx}
\usepackage{textcomp}
\usepackage{xcolor}

\usepackage[comma,numbers,square,sort&compress]{natbib}
\usepackage{epstopdf}
\usepackage{amsmath,amsfonts}
\usepackage{algorithmic}
\usepackage{algorithm}
\usepackage{array}
\usepackage{textcomp}
\usepackage{stfloats}
\usepackage{url}
\usepackage{verbatim}
\usepackage{graphicx}
\usepackage{float}
\usepackage{bm}
\usepackage{multirow}
\usepackage{setspace}
\usepackage{threeparttable}
\usepackage{silence}
\usepackage{amssymb}
\usepackage{color}
\usepackage{xcolor}
\usepackage{pifont}
\usepackage{utfsym}
\usepackage{tabularx}
\usepackage{siunitx}
\UseRawInputEncoding
\usepackage{float}
\usepackage[caption=false,font=footnotesize,labelfont=rm,textfont=rm]{subfig}

\def\BibTeX{{\rm B\kern-.05em{\sc i\kern-.025em b}\kern-.08em
	T\kern-.1667em\lower.7ex\hbox{E}\kern-.125emX}}
\begin{document}

\title{A Learning-based Planning and Control Framework for Inertia Drift Vehicles \\
	
\thanks{*This work is supported by Jianbing Lingyan Foundation of Zhejiang Province, P.R. China (Grant No. 2023C01022) and Major Project of Science and Technology of Yunnan Province, China under Grant 202402AD080001.}
}

\author{\IEEEauthorblockN{ Bei Zhou$^{1}$, Zhouheng Li$^{1}$, Lei Xie*$^{1}$, Hongye Su$^{1}$, Johannes Betz$^{2}$}
	\thanks{*Corresponding Author}
	\thanks{$^{1}$Bei Zhou, Zhouheng Li, Lei Xie, Hongye Su are with the State Key Laboratory of Industrial Control Technology, Zhejiang University, 310058 Hang zhou, China
		{\tt\small zhoubei@zju.edu.cn, zh.li@zju.edu.cn, lxie@iipc.zju.edu.cn, hysu@iipc.zju.edu.cn}}%
	\thanks{$^{2}$Johannes Betz is with the Professorship of Autonomous Vehicle Systems, Technical University of Munich, 85748 Garching, Germany; Munich Institute of Robotics and Machine Intelligence (MIRMI).
		{\tt\small johannes.betz@tum.de}}%
}
\maketitle

\begin{abstract}
	Inertia drift is a transitional maneuver between two sustained drift stages in opposite directions, which provides valuable insights for navigating consecutive sharp corners for autonomous racing.
	However, this can be a challenging scenario for the drift controller to handle rapid transitions between opposing sideslip angles while maintaining accurate path tracking. 
	Moreover, accurate drift control depends on a high-fidelity vehicle model to derive drift equilibrium points and predict vehicle states, but this is often compromised by the strongly coupled longitudinal-lateral drift dynamics and unpredictable environmental variations.
	To address these challenges, this paper proposes a learning-based planning and control framework utilizing Bayesian optimization (BO), which develops a planning logic to ensure a smooth transition and minimal velocity loss between inertia and sustained drift phases. 
	BO is further employed to learn a performance-driven control policy that mitigates modeling errors for enhanced system performance.
	Simulation results on an 8-shape reference path demonstrate that the proposed framework can achieve smooth and stable inertia drift through sharp corners.
\end{abstract}

\begin{IEEEkeywords}
	inertial drift, Bayesian optimization, learning-based control.
\end{IEEEkeywords}

\section{Introduction}

Autonomous drifting is a professional driving skill using a combination of throttle, brakes, and steering inputs to acquire high-sideslip angle maneuvers \cite{AutonomousDrifting2025, AutonomousVehicles2022a}. 
Precise planning and execution of autonomous drifting can significantly extend a vehicle’s maneuverability beyond conventional driving limits. 
Research in drifting can be broadly categorized into three types: sustained drift, which focuses on maintaining the vehicle's state around a set of equilibrium points \cite{LearningBasedHierarchical2024}; transient drift, which involves temporarily entering drift states to perform specific tasks like drift parking or collision avoidance \cite{StableHandling2024, CollisionAvoidance2021}; and inertia drift, which connects two sustained drift maneuvers in opposite directions \cite{RealTimeDriftDriving2022}.
Recently, inertia drift has become a hot research topic, as it provides valuable insights for navigating consecutive sharp corners with minimal speed loss.

The key issue for inertia drift control is determining the optimal transition time between sustained drift and inertia drift maneuvers. 
A well-executed inertia drift should ensure the vehicle smoothly exits one sustained drift and sets up favorable initial conditions for the subsequent sustained drift maneuver, while maintaining controllable and stable drift performance.
Lu et al. \cite{ConsecutiveInertia2023b} proposed a primitive-based planning approach for inertia drift vehicles, which involves additional analysis to generate a look-up table for the inertia drift maneuver. 
However, the planning results often lead to significant curvature and velocity loss, which can negatively impact vehicle performance, especially in racing scenarios with sharp, opposite turning directions.

Inertia drift also presents a more challenging scenario for the drift controller, as it must handle opposite sideslip angle transitions, while maintaining good path tracking performance to smoothly connect two sustained drift maneuvers.
The discovery of unstable drift equilibrium points provides a solid foundation for the theoretical analysis of inertia drift maneuvers \cite{VehicleDrifting2022}.
These equilibrium points are derived from the nominal vehicle dynamics, and assumed as desired vehicle states by the linear quadratic regulator or model predictive control (MPC) \cite{DynamicDrifting2023}.  
However, accurate calculation of these drift equilibrium points can be challenging due to the highly nonlinear vehicle motion equations and varying environmental conditions.

To solve the problem of model mismatch, Weber et al. \cite{ModelingControl2024a} incorporated weight transfer and wheel speed to develop a novel vehicle model, but the extended vehicle states will exacerbate the underactuated control challenges for drift vehicles.
Besides, learning-based approaches, such as diffusion models \cite{OneModel2024} and Gaussian processes \cite{SparseGaussian2025}, are utilized to learn data-driven dynamics models for enhanced control performance.
However, these approaches rely on large amounts of offline trajectory data for training, which can be time-consuming and limit their practicality in real-time applications.

To address the challenges of modeling mismatch and determining the optimal transition time, a learning-based planning and control approach based on Bayesian optimization (BO) is proposed in this paper.
BO is a black-box optimization approach with high-sample efficiency, which is well-suited to learn parameters in expensive-to-evaluate objective functions \cite{TakingHuman2016}.
In the proposed framework, BO is utilized to learn a trigger condition to initialize the inertia drift phase, meanwhile handling model mismatch to ensure feasible and effective control performance.

The main contributions of this paper are as follows:
\begin{enumerate}
	
	\item A learning-based planning and control framework based on BO is proposed to learn the planning logic between inertia drift and sustained drift stages, meanwhile maintaining good drifting and path tracking performance. 	
	
	\item The inertia drift planning logic is developed by learning a trigger condition to eliminate the need for predefined inertia drift primitives, which ensures smooth transitions and minimal velocity loss during inertia drifting. 
	
	\item A performance-driven drift control strategy is introduced to handle modeling mismatch by identifying the reference drift velocity and establishing a feedback control law to provide real-time adjustments for enhanced system performance.

	\item Simulation results on the Matlab-Carsim platform demonstrate that the proposed framework enables smooth and stable inertial drift performance on an 8-shape path with sharp corners.
	
\end{enumerate}

\section{Preliminaries}

\subsection{Drift Vehicle Dynamics}

\begin{figure}[h!]
	\centering
	\includegraphics[width=0.95\linewidth]{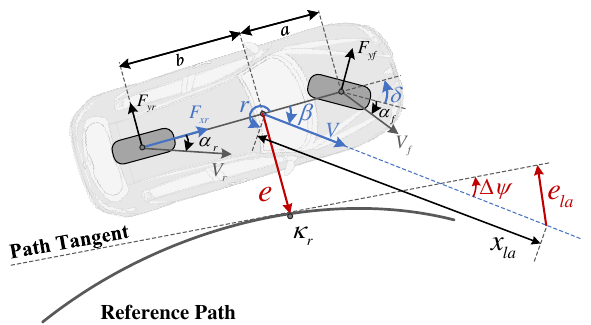}
	\caption{The drift vehicle model.}
	\label{drift_model}
\end{figure}

In this paper, a three-state vehicle model is utilized to describe drifting dynamics as follows:
\begin{align} 
	\dot{V} &= \frac{-F_{yf}\sin{(\delta-\beta)} + F_{yr}\sin{\beta} + F_{xr}\cos{\beta}}{m} \\
	\dot{\beta} &= \frac{F_{yf}\cos{(\delta-\beta)} + F_{yr}\cos{\beta} - F_{xr}\sin{\beta}}{mV} -r \\
	\dot{r} &= \frac{aF_{yf}\cos{\delta} - bF_{yr}}{I_z}
\end{align}
where $V$, $\beta$, $r$ and $\delta$ denote the longitudinal speed, sideslip angle, yaw rate, and steering angle at the center of gravity (CoG), as illustrated in Fig. \ref{drift_model}.
$a$ and $b$ denote the distances from the CoG to the front and rear axles.
The lateral tire forces on the front and rear wheels $F_{yf}$ and $F_{yr}$ are modeled using a simplified Pacejka tire formula model \cite{TyreModelling1987} as follows: 
\begin{align}
	F_{yi} &= -\mu F_{zi}\sin(C\arctan(B\alpha_{i}))  
	\label{Fyi}
\end{align}
where $\mu$ represents the friction coefficient on the ground, $F_{yi}$ and $ F_{zi}$ $(i = f, r)$ represent the lateral tire forces and vertical loads on the front and rear tires.
$B$ and $C$ are tire model coefficients, and $ \alpha_{i}$ $(i = f, r)$ represents the tire sideslip angles for the front and rear tires, which are formulated as: 
\begin{align} 
	\alpha_{f} &= \tan^{-1}(\frac{V\sin\beta + ar }{V\cos\beta}) - \delta\\
	\alpha_{r} &= \tan^{-1}(\frac{V\sin\beta - br }{V\cos\beta}) 
\end{align}

\subsection{MPC Drift Controller}
MPC is a widely used vehicle control strategy that optimizes control actions by predicting future system behavior over a finite horizon.
In this paper, a linear MPC-based drift controller is developed to generate effective control inputs for maintaining the vehicle in deep drift states.

For sustained drift vehicles, the control objective is to stabilize the vehicle states around the drift equilibria, which are acquired by setting the derivatives of vehicle states to zero as $\dot{V}$ = $\dot{\beta} $ = $\dot{r}$ = 0  \cite{ControllerFramework2014a}.
Subsequently, the vehicle model is linearized around these equilibria and discretized using the forward-Euler method to formulate a quadratic programming problem formulation, with more details in our previous work \cite{AdaptiveLearningbased2025b}.
The cost function of the MPC drift controller is formulated as follows:
\begin{align} 
	\text{min}\sum\limits_{k=1}^{N_p} ||\bm { \xi}_k- \bm \xi^{eq}  ||^2_{\bm Q} + \sum\limits_{k=1}^{N_c} ||\Delta \bm \tau_k||^2_{\bm R} 
\end{align} 
\begin{align*} 
	\text{s.t.} \quad
	\bm {\xi}_{k+1} &= f(\bm\xi_k, \Delta \bm \tau_k) \\
	\bm \tau  \in \mathcal{T}, & \ \Delta \bm \tau \in \mathcal{W} 
\end{align*} 
where $ \bm \xi_{k} = [ V, \beta, r, \delta,  F_{xr} ] $ denotes the drift vehicle states, $ \bm \xi^{eq} = [ V^{eq}, \beta^{eq}, r^{eq}, \delta^{eq}, F_{xr}^{eq}] $ denotes the drift equilibria derived from the system model.
The control inputs are defined as $\bm \tau = [\delta, F_{xr}]$, and $\Delta \bm \tau = [\Delta \delta, \Delta F_{xr}]$ the changes between consecutive control actions.
$N_p$ and $N_c$ denote the predictive and control horizons, while $\mathbf{Q} $ and $\mathbf{R} $ are weighting matrices.
The control constraints are defined as $\bm \tau \in \mathcal{T}: = \{|\delta| \leq \delta_{\text{th}}, F_{\text{min}} \leq F_{xr} \leq F_{\text{max}} \}$ and $\Delta \bm \tau \in  \mathcal{W}:  \{ |\Delta \delta| \le \Delta \delta_{\text{th}}, |\Delta F_{xr}| \le \Delta F_{\text{th}} \}$.

This objective function design promotes stable drift behavior and safe driving, facilitating smooth transitions between different drift maneuvers.
However, the inherent modeling mismatch in drift dynamics can pose a significant challenge to accurately calculate the drift equilibria.
To address this problem, a learning-based planning and control framework is implemented to refine the nominal drift equilibria, with more details in the next section.

\begin{figure*}[t!]
	\centering
	\includegraphics[width=1\linewidth]{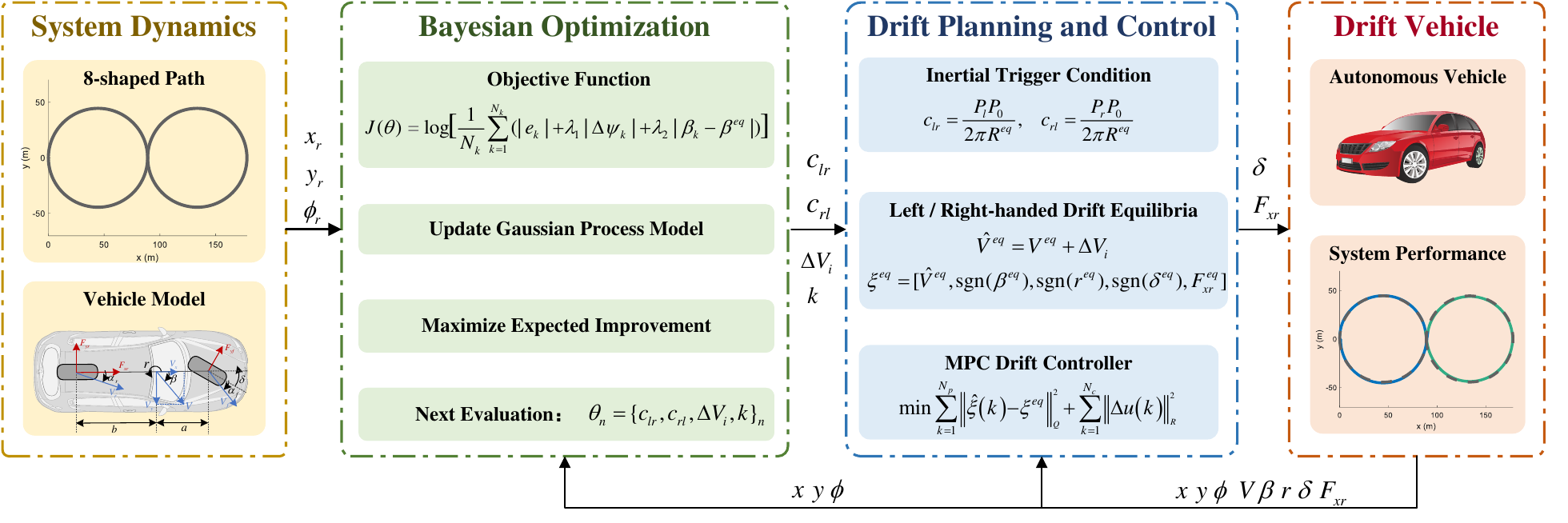}
	\caption{The proposed learning-based planning and control framework.}
	\label{BO_system}
\end{figure*}

\section{The proposed Learning-based Planning and Control Framework}
Inertia drifting depends on two key factors: planning the optimal switch time between sustained and inertia drift phases, and executing precise drift control to effectively manage different drift maneuvers.
In this section, we propose a learning-based planning and control framework to address these critical objectives for desired autonomous drift performance.

\subsection{Inertia Drift Planner}

The objective of inertia drifting is to ensure a smooth and controlled transition between two sustained drift states with opposite driving directions. 
High-quality inertia drifting requires determining the optimal timing for switching between these stages to achieve smooth transitions and minimal deviation from the desired path. 
In this paper, we propose an inertia drift planning logic by learning a trigger condition to decide the optimal switch time between different drift strategies.

\subsubsection{Planning Logic}

Instead of relying on a pre-stored library of inertia drift primitives, our approach directly uses the switched sustained drift states as reference points for the inertia drift behavior.
For example, the vehicle first performs a left-handed drift with the reference states $\xi^{eq}_{l}$, transitions into inertia drift, and then enters a right-handed drift with the reference states $\xi^{eq}_{r}$.
The inertia drift phase begins with the trigger condition, with the same reference states as the following right-handed drift. 
This approach avoids redundant analysis of inertia drift behavior, enabling swift transitions with minimal velocity loss.

\subsubsection{Trigger Condition}

\begin{figure}[h!]
	\centering
	\includegraphics[width=0.95\linewidth]{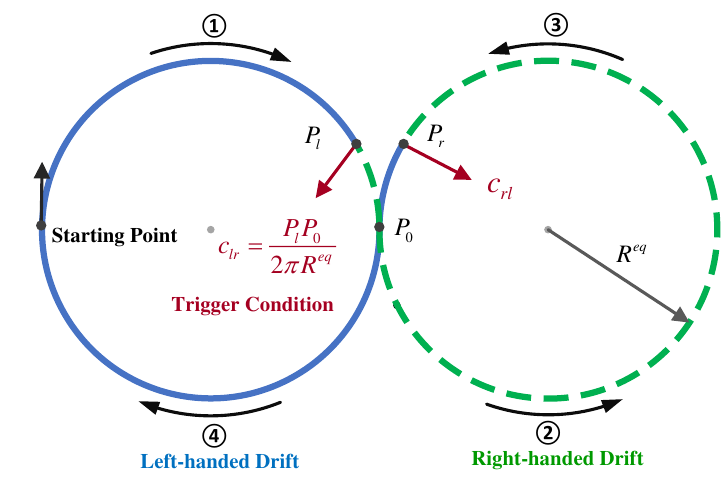}
	\caption{Inertia drift planning on an 8-shaped path.}
	\label{inertial}
\end{figure}

There are two trigger conditions to be learned: the first, $c_{lr}$, corresponds to the vehicle exiting its current sustained drift to enter the inertia drift phase; the second, $c_{rl}$, relates to transitioning from inertia drift to another sustained drift state.
Both trigger conditions are designed with the vehicle's current position on the reference path as:  
\begin{align} 
	c_{lr} = \frac{\overset{\frown}{P_lP_0}}{2\pi R^{eq}}, \quad
	c_{rl} = \frac{\overset{\frown}{P_rP_0}}{2\pi R^{eq}}
\end{align}
where $P_l(x_l, y_l)$ and $P_r(x_r, y_r)$ denote the closest positions on the left-handed and right-handed sustained drifting paths to the vehicle’s current position, $P_0(x_0, y_0)$ represents the intersection point of two sustained drift paths, $\overset{\frown}{P_lP_0}$ and $\overset{\frown}{P_rP_0}$ represent the arc length between two points, $R^{eq}$ is the drift radius of the sustained drift path. 
Once the trigger condition is met, the reference drift equilibrium points are updated accordingly.
This design allows for a general planning logic that can be applied to various sustained drift states with different drift radii, as illustrated in Fig. \ref{inertial}.

\subsection{ Performance-driven Drift Control}
The complexity of drifting dynamics necessitates a precise control strategy to keep the vehicle states close to the desired drift equilibrium points while adhering to the given path. 
In this paper, we propose a performance-driven drift control strategy that works in conjunction with the MPC drift controller to enhance system performance.

\subsubsection{Identify Desired Drift Velocity}
The effectiveness of MPC control largely depends on the accuracy of drift equilibrium points $\xi^{eq}$ to serve as control targets.
However, due to environmental changes and coupled longitudinal-lateral dynamics, it can be challenging to analytically derive precise drift equilibrium points.
Additionally, the MPC drift controller incorporates a linearization process for state prediction, which can introduce further inaccuracies and degrade control performance. 
To address this, a system identification process is applied to compensate for modeling errors caused by highly nonlinear vehicle dynamics and uncertain environmental factors.

The identification process involves developing system dynamics from experimental data to facilitate effective controller design. The two sustained drift states, connected by inertia drift, have opposite sideslip angles, yaw rates, and steering angles as their drift equilibrium points. These relationships can be represented as:
\begin{align} 
	\bm \xi^{eq} = [ V^{eq}, \text{sgn}(\beta^{eq}), \text{sgn}(r^{eq}), \text{sgn}(\delta^{eq}), F_{xr}^{eq}]
\end{align} 
with the sign function defined as:
$$
\text{sgn}(x) =
\begin{cases} 
	-1, & \text{left-handed drift}, \\
	1, & \text{right-handed drift}.
\end{cases}
$$
It can be observed that these two desired sustained drift states share the same equilibrium points for velocity and tire force, with other states having the same values but opposite directions.
Motivated by this, we propose a unified identification process that correlates these two sustained drifting phases by learning a residual term $\Delta V_i$, which modifies the desired drift velocity as:
\begin{align} 
	\hat V^{eq} = V^{eq} + \Delta V_i
\end{align} 
where $V^{eq}$ is the drift velocity derived from system model and $\hat V^{eq}$ is the desired control target for the MPC controller.

\subsubsection{ Feedback Control for Path Tracking}
For drifting vehicles, precise path tracking is a prerequisite for the successful transition between different drift maneuvers. 
Since the tracking error during sustained drift tends to grow more pronounced, especially after switching drift strategy with rapid change in drift equilibrium points.
To address this issue, a feedback control based on an expert strategy is proposed to provide real-time adjustments for enhanced path tracking performance.

The expert strategy is utilized by professional drivers to dynamically adjust the steering control to follow the desired racing line.
Inspired by this, a relationship between the steering angle and the tracking error can be established as: 
\begin{align} 
	\delta_f = \delta + k\cdot e_{la}
\end{align} 
where $\delta^{eq}$ is the drift equilibrium point, $\delta$ is the steering angle computed by the MPC drift controller, $k$ is a proportional gain parameter to be learned, and $e_{la}$ is the lookahead error to anticipate the lateral deviation projected at a look-ahead distance  $x_{la}$, which can be calculated by $e_{la} = e + x_{la} \cdot\sin (\Delta{\psi})$.
The heuristics for the designed feedback control law can refer to our previous paper  \cite{LearningBasedHierarchical2024}.
A learning-based approach is then applied to provide precise quantitative analysis to establish the control feedback law by learning the optimal parameter $k$.

\begin{algorithm}[!b]
	\caption{Learning-Based Framework with BO}
	\label{BO}
	\setstretch{1}
	\begin{algorithmic}[1]
		
		\STATE Initialize the GP model
		\FOR{iteration $n = 1$ to $N$}
		
		\STATE Get $\bm{\theta}_n = \{c_{lr}, c_{rl}, \Delta V_i, k\}_n$ to be evaluated
		\STATE Set desired drift velocity: $\hat V^{eq} = V^{eq} + \Delta V_i$
		
		\IF{trigger condition $c_{lr}$ is satisfied}
		\STATE Switch into right-handed sustained drift with: $\xi_r^{eq}$
		\ELSIF{trigger condition $c_{rl}$ is satisfied}
		\STATE Switch into left-handed sustained drift with: $\xi_l^{eq}$
		\ELSE 
		\STATE Begin left-handed sustained drift with $\xi_l^{eq}$
		\ENDIF
		
		\STATE Compute drift vehicle inputs from MPC: $u = [\delta, F_{xr}]$
		\STATE Apply feedback control law: $\delta_f = \delta + k \cdot e_{la}$
		
		\ENDFOR
		\STATE \textbf{Output:} Optimized parameters $\bm{\theta}^* = \{c_{lr}^*, c_{rl}^*, \Delta V_i^*, k^*\}$
		
	\end{algorithmic}
\end{algorithm}

\begin{figure}[!b]
	\centering
	\subfloat[Velocity]{\includegraphics[width=1\linewidth]{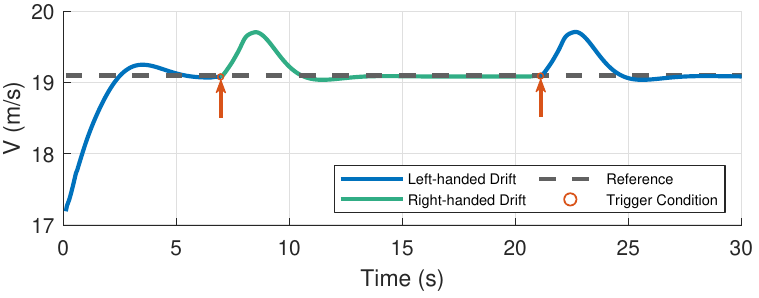}%
		\label{v}}
	\hfil
	\subfloat[Sideslip angle]{\includegraphics[width=1\linewidth]{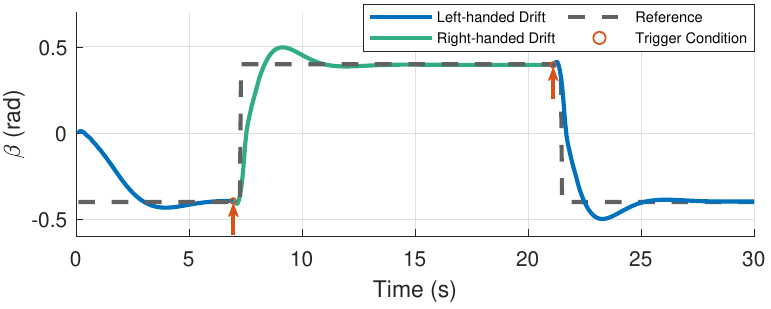}%
		\label{beta}}
	\hfil
	\subfloat[Yaw rate]{\includegraphics[width=1\linewidth]{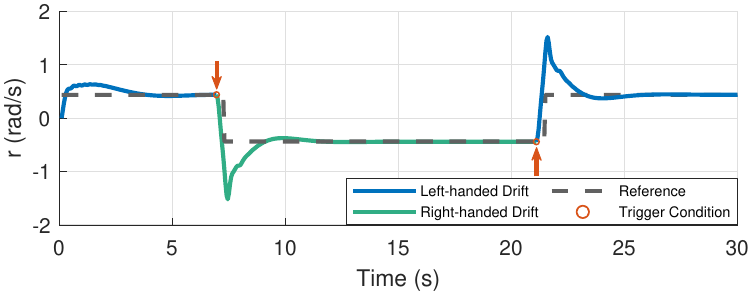}%
		\label{r}}
	
	\hfil
	\centering
	\subfloat[Steering angle]{\includegraphics[width=1\linewidth]{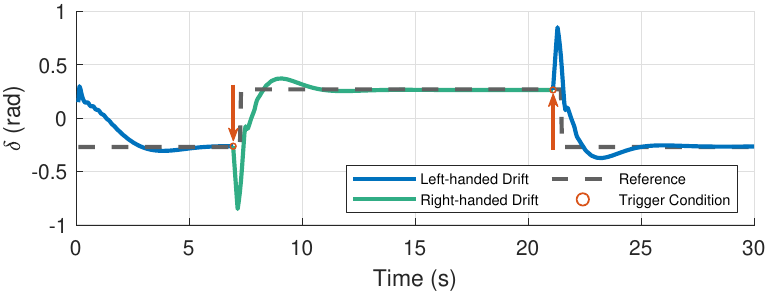}%
		\label{delta}}
	
	\caption{Simulations results of drift vehicle states.} 
	
	\label{states}
\end{figure}

\subsection{Learning-based Framework with Bayesian Optimization}

BO is a sample-efficient algorithm designed to acquire optimal system parameters by minimizing expensive-to-evaluate objective functions.
It builds a surrogate probabilistic model to approximate the unknown objective function based on data collected from previous evaluations.  
Then an acquisition function balancing exploration and exploitation is utilized to decide the next evaluation point.
In this paper, we choose the Gaussian Process as the surrogate model and expected improvement as the acquisition function to construct the optimization problem as: 
\begin{align} 
	\bm{\theta^*} &= \arg\min_{\bm{\theta \in \Theta}} J(\bm{\theta})
\end{align}	
where $\bm{\theta^*}$ denotes the optimal parameter set to be learned, and $J(\bm{\theta})$ is the evaluate function.
By continuously acquiring new data and updating the GP model, BO progressively converges toward the global optimum.

For inertial drift vehicles, the parameters to be learned are defined as $\bm{\theta} = [ c_{lr}, c_{rl}, \Delta V_i, k] \in \mathbb{R}^4$.
The optimization process of BO is guided by enhancing drift performance through minimizing the sideslip angle error, and improving path tracking performance by minimizing the lateral error $e_k$ and the course direction error $\Delta \psi_k$.
The optimization objective is formulated as:
\begin{align} 
	J(\bm{\theta}) &= \text{log}  \big[ \frac{1}{N_k} \sum\limits_{k=1}^{N_k} (|e_k| + \lambda_1|\Delta \psi_k| + \lambda_2|\beta_k - \beta^{eq}|)  \big]
\end{align} 
where $N_k$ is the total number of timesteps during an iteration, $\lambda_1$ and $\lambda_2$ are two constant weighting factors to balance the vehicle performance. 
The sideslip angle $\beta$ is the most critical state for characterizing drift maneuvers, while the regulation of other drift states is primarily handled by the MPC controller.
The BO learning process is summarized in Algorithm \ref{BO}.

\section{Simulation }
In this section, we validate the proposed framework on the Matlab-Carsim platform to demonstrate its effectiveness in achieving smooth transitions between inertial drift and sustained drift maneuvers.

\subsection{Simulation Setup}

The reference path used in the simulation is an 8-shaped track composed of two tangent circles, each with a radius of 44 m. 
The vehicle begins with a left-handed sustained drift maneuver for half a circle, then transitions into inertia drift to smoothly switch into a right-handed sustained drift maneuver. 
After completing a full lap of the right-handed sustained drift, the vehicle performs another inertia drift to transition back into a left-handed sustained drift for another half lap.

The sampling time of the MPC drift controller is set to $\Delta T = 0.05$ s, with both the prediction horizon $N_p$ and control horizon $N_c$ set to 20. 
BO is trained for 500 iterations, with each objective function evaluation lasting $T = 30$ s for a whole 8-shaped path, and 20 initial seed points used to initialize the GP model. 
The active set size for the GP model is set to 400 to balance model accuracy and computational efficiency. 
The weighting parameters in the BO objective function are set as $\lambda_1 = 5$ and $\lambda_2 = 1$.
Additional system parameters are provided in our previous work \cite{AdaptiveLearningbased2025b}.

\subsection{Simulation Results}

The sustained drift equilibria derived from the vehicle model and BO-optimized parameters are illustrated in Table \ref{para}. 
The vehicle states along the entire path are illustrated in Fig. \ref{states}, where the reference states are related to the predefined tangent circles.
The blue and green lines represent the actual drift behavior executed by the MPC drift controller. While the orange markers indicate the trigger points initiating transitions between left and right-handed sustained drift.

\begin{table}[!h]
	\caption{Online Parameters. \label{para}}
	\centering
	\begin{threeparttable}
		\begin{tabular}{ cc|cc }        
			\hline
			Drift Equilibria  & Value & BO Parameters  & Value  \\
			\hline
			$V^{eq}$ & 19.1 m/s & $c_{lr}$ & 0.0273 \\
			$\beta^{eq}$ & 0.4 rad & $c_{rl}$ & 0.0215   \\
			$r^{eq}$ & 0.434 rad/s  & $\Delta V_{i}$ & 0.12  \\
			$\delta^{eq}$ & 0.27 rad & $k$ & 0.21  \\
			$F_{xr}^{eq}$ & 4324 N 	&  	   & \\
			
			\hline
		\end{tabular}
	\end{threeparttable}
\end{table}

\begin{figure}[!ht]
	\centering
	\subfloat[Path tracking performance in Matlab]{\includegraphics[width=0.95\linewidth]{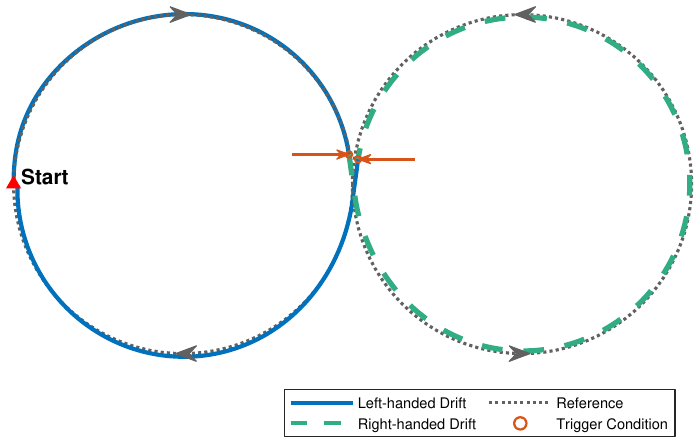}%

		\label{matlab}}
	\hfil
	\subfloat[Path tracking performance in Carsim ]{\includegraphics[width=0.95\linewidth]{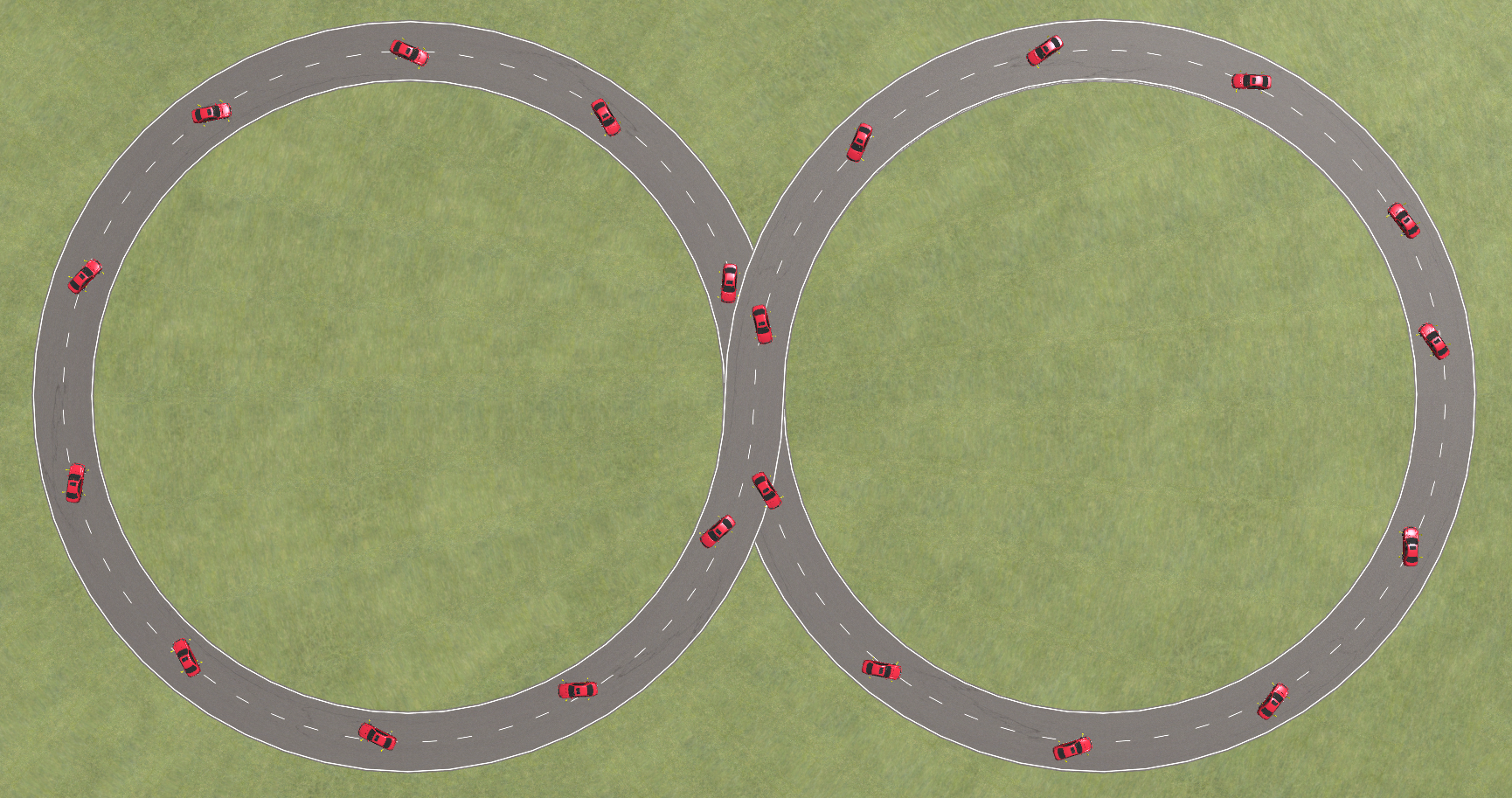}%
		\label{high_view}}
	\hfil
	\subfloat[Inertial drift vehicle]{\includegraphics[width=0.95\linewidth]{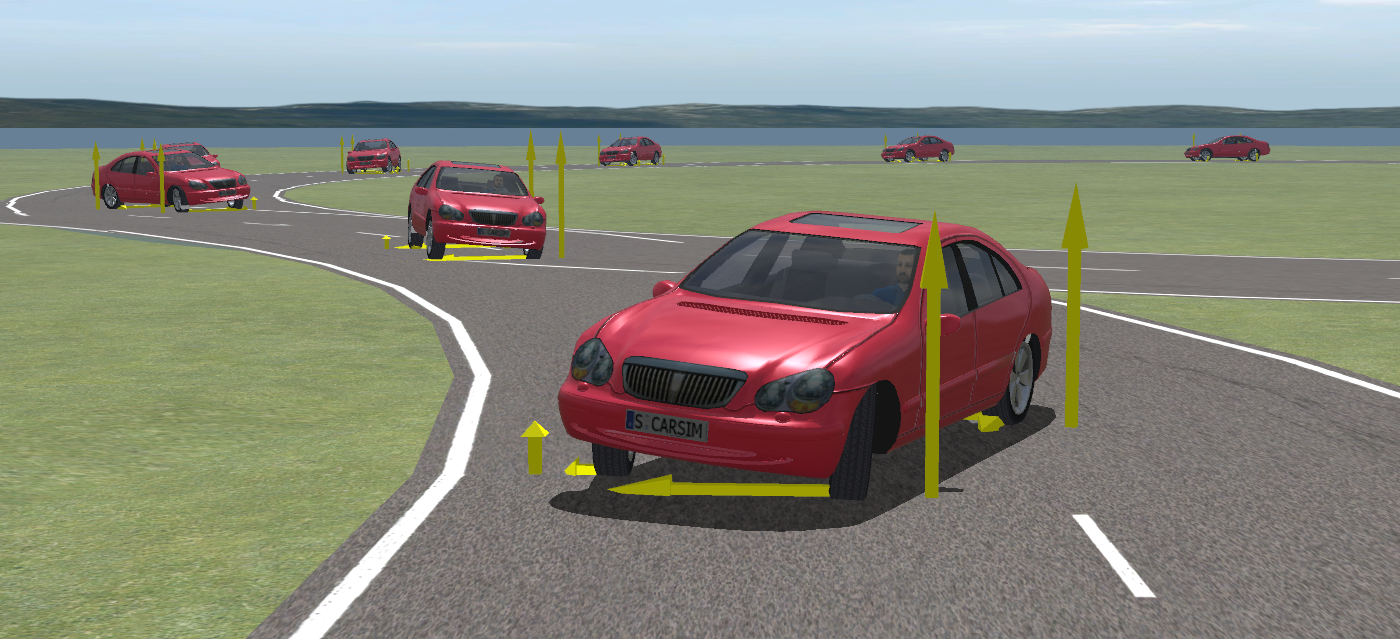}%
		\label{drift}}
	
	\caption{Simulations results on the Matlab-Carsim platform.} 
	
	\label{position}
\end{figure}

It can be observed that the learned trigger condition enables the vehicle to switch drift maneuvers slightly in advance, thereby facilitating a smooth inertia drift transition between opposite sustained drift directions. 
As shown in Fig. \ref{beta}, the sideslip angle rapidly turns into an opposite direction after the trigger points, indicating a successful inertia drift maneuver. 
Although this reduction in sideslip angle leads to a brief increase in velocity (Fig. \ref{v}), the learned control policy provides adaptive adjustments that allow the velocity to rapidly converge to the desired value. 
The yaw rate (Fig. \ref{r}) and steering angle (Fig. \ref{delta}) exhibit temporary deviations from their references during the transition. 
These deviations reflect a short period of normal driving as the vehicle exits the left-handed drift and prepares to enter the right-handed drift, which helps facilitate a stable transition.

The overall 8-shaped drifting simulation in MATLAB is shown in Fig. \ref{matlab} and the Carsim animation of our approach is presented in Fig. \ref{high_view} and Fig. \ref{drift}. 
The results show that the vehicle is always adhere to the given path even during transitions between sustained and inertia drift maneuvers, which validate the effectiveness of the proposed learning-based framework in managing dynamic and nonlinear drift behaviors.

\section{Conclusion}

This paper proposes a learning-based planning and control framework for inertia drift vehicles, capable of achieving smooth and stable transitions between sustained drift maneuvers in opposite directions.
To solve the key challenge of determining the optimal switching time between drift strategies, BO is utilized to learn a trigger condition associated with the completion of a sustained drift.
In addition, BO is employed to mitigate modeling mismatches by identifying the desired drift velocity and a feedback control law for improved system performance.
Simulation results on an 8-shaped path demonstrate that the proposed framework effectively ensures smooth and stable drift transitions with minimal velocity loss.
In future work, we will focus on evaluating the framework’s generalization capability for inertial drift between two sustained drift paths with different radii.

\bibliographystyle{IEEEtran}
\bibliography{conf_reference}

\vspace{12pt}

\end{document}